\documentclass{article}

\PassOptionsToPackage{numbers, compress}{natbib}

\usepackage[preprint]{neurips_2020}

\usepackage[utf8]{inputenc} 
\usepackage[T1]{fontenc}    
\usepackage{hyperref}       
\usepackage{url}            
\usepackage{booktabs}       
\usepackage{amsfonts}       
\usepackage{nicefrac}       
\usepackage{microtype}      

\usepackage{graphicx}
\usepackage{subfigure}
\usepackage{amsmath}
\usepackage{multirow}
\title{Interventional Domain Adaptation}

\author{%
  Jun Wen \\
  College of Computer Science and \\
  Technology, Zhejiang University\\
  \texttt{junwen@zju.edu.cn} \\
   \And
   Changjian Shui \\
   Laval University\\
  \texttt{	changjian.shui.1@ulaval.ca} \\
  \And
  Kun Kuang \\
  College of Computer Science and \\
  Technology, Zhejiang University\\
  \texttt{kunkuang @zju.edu.cn} \\
   \And
  Junsong Yuan \\
  State University of New York at Buffalo \\
  \texttt{jsyuan@buffalo.edu} \\
  \And
  Zenan Huang \\
  College of Computer Science and \\
  Technology, Zhejiang University\\
  \texttt{lccurious@zju.edu.cn} \\
  \And
  Zhefeng Gong \\
  Zhejiang University \\
  School of Medicine \\
  \texttt{zfgong@zju.edu.cn} \\
  \And
  Nenggan Zheng \\
  Qiushi Academy for Advanced \\
  Studies, Zhejiang University\\
  \texttt{zng@cs.zju.edu.cn} \\
}

\begin{document}

\maketitle

\begin{abstract}

Domain adaptation (DA) aims to transfer discriminative features learned from source domain to target domain. Most of DA methods focus on enhancing feature transferability through domain-invariance learning. However, source-learned discriminability itself might be tailored to be biased and unsafely transferable by spurious correlations, \emph{i.e.}, part of source-specific features are correlated with category labels. We find that standard domain-invariance learning suffers from such correlations and incorrectly transfers the source-specifics. To address this issue, we intervene in the learning of feature discriminability using unlabeled target data to guide it to get rid of the domain-specific part and be safely transferable. Concretely, we generate counterfactual features that distinguish the domain-specifics from domain-sharable part through a novel feature intervention strategy. To prevent the residence of domain-specifics, the feature discriminability is trained to be invariant to the mutations in the domain-specifics of counterfactual features. Experimenting on typical \emph{one-to-one} unsupervised domain adaptation and challenging domain-agnostic adaptation tasks, the consistent performance improvements of our method over state-of-the-art approaches validate that the learned discriminative features are more safely transferable and generalize well to novel domains.

\end{abstract}

\section{Introduction}

\emph{Domain shift} is commonly encountered by machine learning practitioners that a model is trained in one specific distribution, denoted as the source domain, but applied to another different yet related distribution, denoted as the target domain. Domain adaptation (DA) is proposed to generalize learning machines and alleviate the performance degradations caused by \emph{domain shift}, which is studied in varied fields, \emph{e.g.}, computer vision \cite{csurka2017domain} and nature language processing \cite{freitag2016fast}.

The key to successful domain adaptation is to learn discriminative features from the source domain that are transferable to the target domain. Feature discriminability is generally learned via supervised training with labeled source data, and feature transferability is most studied and prevalently strengthened by learning domain-invariant features \cite{ben2010theory}. To this end, various DA approaches have been proposed, e,g., maximum mean discrepancy (MMD) minimization  \cite{long2015learning,long2017deep}, moment matching \cite{peng2019moment}, and top-performing adversarial training \cite{ganin2016domain,tzeng2017adversarial}. While transferability is enhanced, the learned feature discriminability might be damaged, as shown in \cite{rozantsev2018beyond,chen2019transferability}, which attempt to maintain source-learned feature discriminability by guiding the target domain to approach the source.

\begin{figure*}[tb]
  \centering
  \subfigure[Colored Office-31]{\includegraphics[scale=0.2]{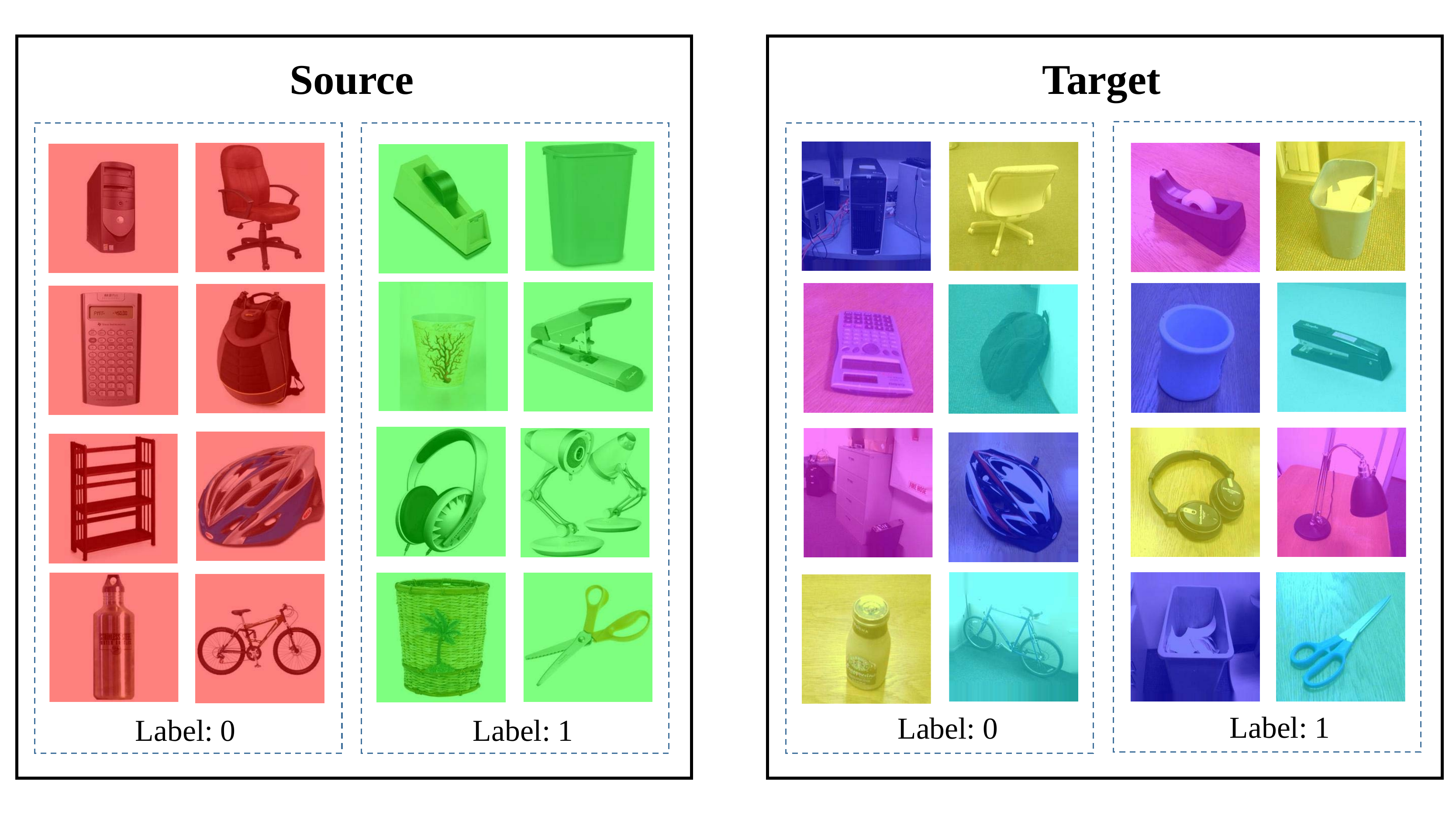}}
  \subfigure[Accuracy]{\includegraphics[scale=0.33]{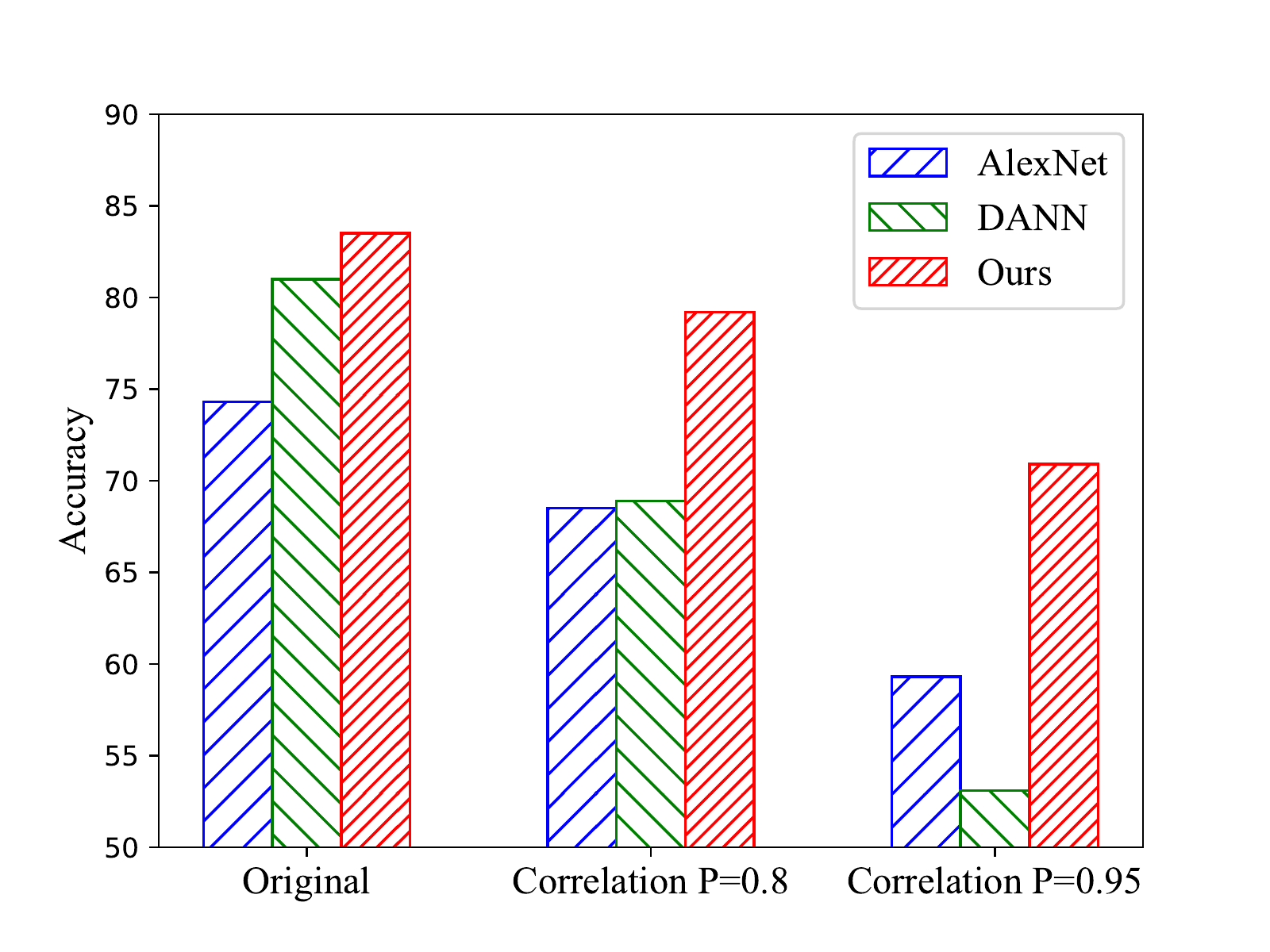}}
  \caption{Domain adaptation results on the \emph{A$\rightarrow$W} binary classification task with colored \emph{Office-31}. In the source data, the category labels are highly correlated with colors. Specifically, category ``0'' is colored by \emph{red} with a probability of $P$ otherwise by \emph{green} and category ``1'' is colored in a reverse way. Higher probability indicates stronger correlation. However, the target data are colored randomly by four different colors. As can be seen, the adversarial domain adaptation, DANN \cite{ganin2016domain}, suffer a significant performance drop when the correlation increases because of the incorrect transfer of ``discriminative'' colors from the source to the target.}
  \label{fig:figure1}
\end{figure*}

However, we find that source-learned feature discriminability itself could be unsafely transferable because of spurious feature-label correlations which is commonly encountered in machine learning practice \cite{arjovsky2019invariant}. To dive into this issue, we perform domain adaptation on a colored \emph{Office-31} dataset, where colors are highly correlated with category labels. As shown in Figure \ref{fig:figure1}, domain adversarial training, DANN \cite{ganin2016domain}, fails to get rid of such spurious correlations and incorrectly transfers the colors as ``discriminative'' features from the source to target. In Figure \ref{fig:figure1}(a), we denote the colors as domain-specific features $V$ whose discriminability is domain-dependent while the object semantics are denoted as domain-sharable features $U$ which are consistently discriminative across domains. $U$ and $V$ are generally unknown in prior and entangled in feature space. Standard domain-invariance learning suffers from such correlations, especially when domain discrepancy in $V$ is more salient than that in $U$, which is common in DA practice, \emph{e.g.}, the dominated variations of image styles across domains in most of the DA testbeds \cite{saenko2010adapting,venkateswara2017deep,peng2019moment}. In this case, standard domain discrepancy minimization is generally dominated by $V$, instead of by $U$, and $V$ is incorrectly transferred by learning a domain-invariant arbitrary mix-up between $P(V_S)$ and $P(V_T)$ with trivial transformations from $V_S$ to $V_T$, which deteriorates the learned feature discriminability and causes the performance drop of DANN by \textbf{27.9\%} in Figure \ref{fig:figure1}(b).

In this paper, we intervene in the learning of feature discriminability with unlabeled target data to attenuate the above spurious correlations by proposing \emph{interventional domain adaptation} (IDA) to learn discriminative features that are safely transferable. Concretely, assuming inputs $X_{S}=(U_S,V_S)$ and $X_{T}=(U_T,V_T)$ from the source $S$ and target $T$, respectively, we first generate counterfactual features $X_{S,T}=(U_S,V_T)$ and $X_{T,S}=(U_T,V_S)$ through a novel feature intervention strategy that distinguishes $U$ from $V$ in entangled feature space via an adversarially-trained feature attention module. Then, to prevent the domain-specifics from residing in the learned feature discriminability, we train both $F_S$ and $F_T$ to be invariant to the mutations of the domain-specifics with the counterfactual features, namely achieving $P(F_S|(U_S,V_S))=P(F_S|(U_S,V_T))= P(F_S|U_S)$ and $P(F_T|(U_T,V_T))=P(F_T|(U_T,V_S))= P(F_T|U_T)$.

The main contributions of our work are summarized as the following. 1) We show that feature discriminability learned from source domain could suffer from spurious feature-label correlations and be unsafely transferable. 2) We propose to intervene in the learning of feature discriminability with unlabeled target data to guide it to focus on domain-sharable part by performing mapping consistency over counterfactual features that are generated through a novel feature intervention strategy. 3) Extensive experimental results on both standard \emph{one-to-one} unsupervised domain adaptation and challenging domain-agnostic adaptation tasks demonstrate the promise of the proposed method by outperforming the state-of-the-art approaches. Further, the learned features are shown to generalize well to novel domains.

\begin{figure*}[]
   \centering
   \includegraphics[width=0.95\linewidth]{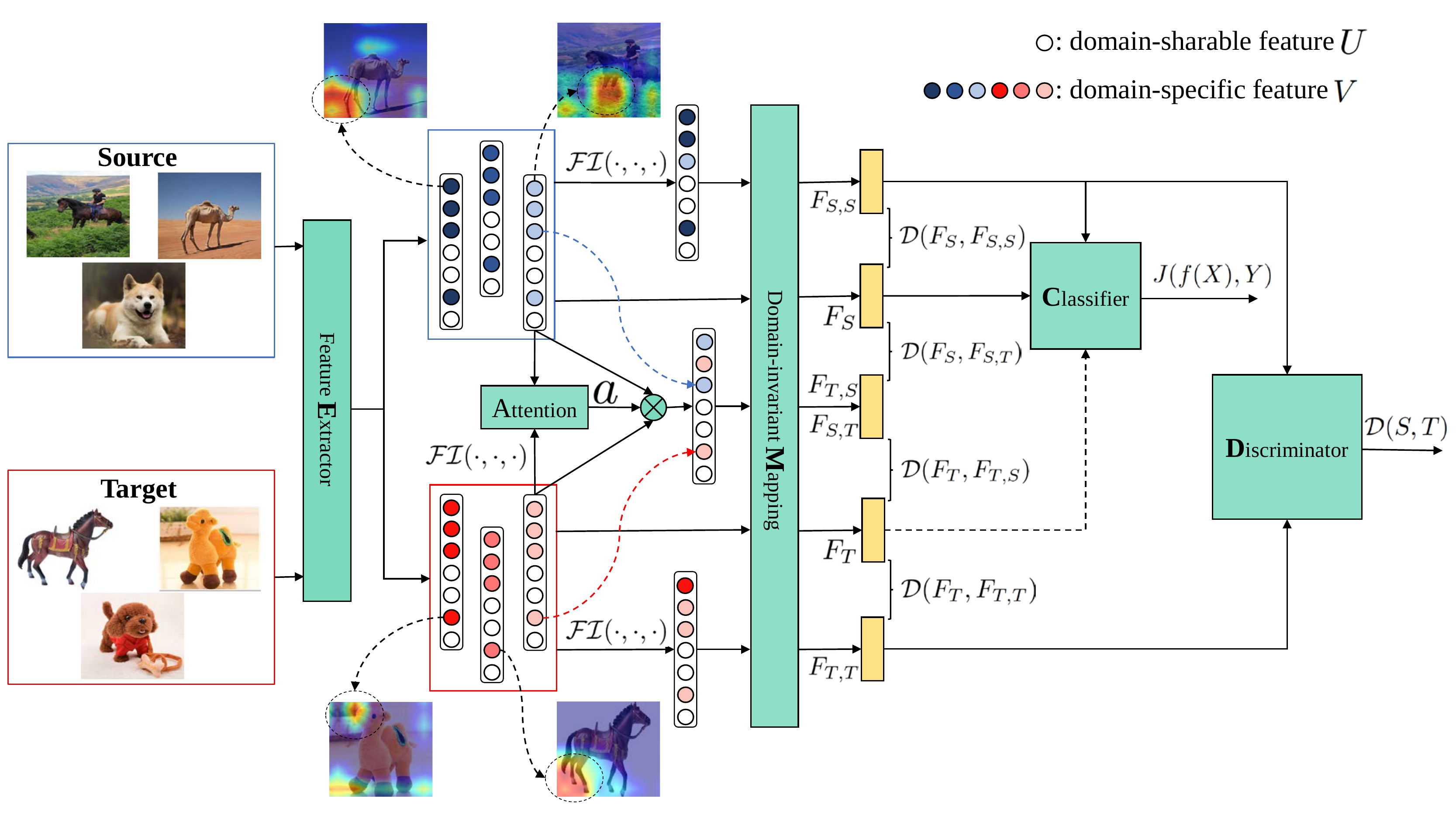}
   \caption{Architecture of the proposed IDA for learning discriminative features that are safely transferable. The parameterized feature intervention operation $\mathcal{FI}(\cdot,\cdot,\cdot)$ is designed to distinguish domain-sharable feature $U$and domain specific $V$. Minimizing $\mathcal{D}(\cdot,\cdot)$ encourages the domain-invariant feature $F$ to focus on $U$ and get rid of $V$.}
\label{fig:pipeline}
\end{figure*}

\section{Related Work}
Domain adaptation aims to transfer learned discriminative features from source to target domain. To strength features transferability, mainstream methods generally learn domain-invariant features \cite{ben2010theory,zhao2019learning}. Among them, the deep adversarial-domain-adaptation approaches are the most popular and have achieved the top performance \cite{ganin2016domain,tzeng2017adversarial,xie2018learning,pei2018multi,saito2018maximum,xu2020adversarial}. However, recent works show that domain-invariance learning could deteriorate the learned feature discriminability and propose various approaches to maintain or strengthen it \cite{rozantsev2018beyond,chen2019transferability,liu2019transferable,kurmi2019attending,chen2020adversarial}. However, in this paper, we show that the source-learned discriminability itself could suffer from source-specific correlations and be of poor transferability, and propose to alleviate the correlations and learn discriminative features that are safely transferable.

\section{Interventional Domain Adaptation}
We present the proposed method for unsupervised domain adaption. We are given a labeled source-domain dataset $D_S=(X_S,Y_S)$ and an unlabeled target-domain dataset $D_T=(X_T$). The $D_S$ and $D_T$ are assumed to be sampled from two different but related distributions $P_S(X_S)$ and $P_T(X_T)$, respectively, with $P_S \ne P_T$. The ultimate goal is to transfer discriminative features learned from $D_S$ to $D_T$ to minimize the label prediction risk on $D_T$. Due to the commonly encountered data bias, DA methods could suffer from the spurious correlations between source-specific feature $V$ and label $Y$, and incorrectly transfer $V$ as important predictors. In this paper, we attenuate the above unsafely transferable correlations and guide the model to focus on the transfer of domain-sharable $U$ by learning domain-invariant feature $F$ which satisfies $P(F|(U,V))= P(F|U)$. 

The architecture of the proposed method is mainly composed of three parts, as shown in Figure \ref{fig:pipeline}. The feature extractor \emph{E} extracts feature $X$, the domain-invariant mapping \emph{M} transforms $X$ into domain-invariant representation $F$, and the classifier \emph{C} predicts label $Y$ upon $F$. We train the network with the following three objectives:
\begin{equation}
        \mathcal{L} = J(f(X),Y)+\beta \mathcal{L}_{FI} +\gamma \mathcal{D}(S,T) ,
\end{equation}
\noindent where $J(f(X),Y)$ is source supervised training objective with a typical cross-entropy loss for feature discriminability. $\mathcal{L}_{FI}$ denotes feature intervention loss which guides the discriminability learning to get rid of domain-specific features and be safely transferable. $\mathcal{D}(S,T)$ measures domain discrepancy. The $\beta$ and $\gamma$ are balance parameters. The detailed training procedure is provided in the supplementary materials. In the following sections, we first describe the feature intervention strategy for the $\mathcal{L}_{FI}$, and then present the $\mathcal{D}(S,T)$ over the counterfactual features.

\subsection{Intervening in Discriminability}

In this section, we explain how to intervene in the learning of feature discriminability and guide it to focus on domain-sharable features. We first present a novel feature intervention strategy to generate counterfactual features. Then, we perform mapping consistency over the counterfactual features to encourage the learned discriminability to be independent of the domain-specifics.

\subsubsection{Counterfactual Feature Generation}
We generate counterfactual features through attention-guided feature intervention. With inputs $X_S=(U_S,V_S)$ and $X_T=(U_T,V_T)$ from the source $S$ and target $T$, respectively, we learn discriminative representation $F_S$ and $F_T$ via supervised training with available source labeled data. To avoid $V$ being learned as important predictors for incorrect transfer, $F$ is guided to be conditionally independent of $V$, given $U$, namely satisfying $P(F_S|(U_S,V_S))=P(F_S|(U_S,V_T))$ and $P(F_T|(U_T,V_T))=P(F_T|(U_T,V_S))$ in which both $X_{S,T}=(U_S,V_T)$ and $X_{T,S}=(U_T,V_S)$ are counterfactual features. To generate them, we devise the following feature intervention strategy $\mathcal{FI}(\cdot,\cdot,\cdot)$:
\begin{equation}
\tilde{x}=\mathcal{FI}(x^{y_a}_h,x^{y_b}_l,W)=  a \cdot x^{y_a}_h+(1-a) \cdot x^{y_b}_l,
\end{equation}
\noindent where $x^{y_a}_h$ and $x^{y_b}_l$, parents of the counterfactual feature $\tilde{x}$, denote $n$-dimensional features with category label $y_a$ and $y_b$, and domain label $h$ and $l$, respectively. The intervention switch value $a=\sigma(x^{y_a}_h,x^{y_b}_l,W)$ is calculated by an attention module \cite{hu2018squeeze} that is parameterized by $W$ and with a \emph{sigmoid} activation function $\sigma(\cdot)$. The attention module determines which part of the counterfactual features are going to be inherited from the two parent features, respectively. It is trained to distinguish $V$ from $U$ and guide the learning of $F$ in an adversarial manner against the \emph{M}.

To be noted, we expect counterfactual features that are ideally with fully inherited or invariant $U$ across domains but variant $V$. However, the feature intervention could incorrectly mix up $U$ and then its influence on F across domains. To alleviate this, the feature intervention is limited only within each category, where $U$ is expected to be invariant. Further, we train the whole model with the source-generated counterfactual feature $\mathcal{FI}(x^{y_a}_S,x^{y_a}_S,W)$ in a supervised manner which is of the same category label $y_a$ to its parents. Such training encourages the whole model to be insensitive to the feature interventions on $U$ within each category.


\subsubsection{Consistent Mapping}
We guide the discriminability learning to focus on the domain-sharable $U$ and get rid of the domain-specific $V$ by optimizing the learned representation $F$ to be invariant to the cross-domain mutations of $V$ when conditioned on $U$. To this end, we first formulate the following distance measure $\mathcal{D}(\cdot,\cdot)$ over $F$:
\begin{equation}
\small
 \mathcal{D}(F_{x^{y_a}_h},F_{x^{y_b}_l})=
\begin{cases}
||F_{x^{y_a}_h}-F_{x^{y_b}_l}||_q &  {y_a} = {y_b}; \\
\max(0,t_d-||F_{x^{y_a}_h}-F_{x^{y_b}_l}||_q) &  {y_a} \neq {y_b},
\end{cases}
\label{eq:distance_define}
\end{equation}
\noindent where $F_{x^{y_a}_h}$ ( or $F_{x^{y_b}_l}$) denotes the representation of sample $x^{y_a}_h$ ( or $x^{y_b}_l$) with label ${y_a}$ ( or ${y_b}$) from domain $h$ ( or $l$). The $t_d$ is distance threshold for samples from different categories. Minimizing the distance encourages samples from the same category to be mapped to consistent $F$ while samples from different categories to distinct $F$.

To achieve $F$'s independence of $V$, we perform mapping consistency over the counterfactual features. Concretely, counterfactual features that are composed of the same domain-sharable $U$ but different domain-specific $V$, namely $(U, V_S)$ and $(U, V_T)$, should be mapped by the \emph{M} into consistent or identical $F$, with the $V_s$ and $V_t$ assumed to be distinct across domains. To this end, we optimize $F$ to minimize the following objective:
\begin{equation}
\small
\mathcal{L}_{FI}=   Q_{x_S}[\mathcal{D}(F_S,F_{S,S})+\mathcal{D}(F_S,F_{S,T})] + Q_{x_T}[\mathcal{D}(F_T,F_{T,S})+\mathcal{D}(F_T,F_{T,T})],
\label{eq:distance_total}
\end{equation}
\noindent where $F_{S,S}=F_{\mathcal{FI}(x_S,x_S,W)}$ and $F_{T,T}=F_{\mathcal{FI}(x_T,x_T,W)}$ denote the representations of the source and target counterfactual features $\mathcal{FI}(x_S,x_S,W)$ and $\mathcal{FI}(x_T,x_T,W)$, respectively. $F_{S,T}$ and $F_{T,S}$ are the representations of the cross-domain counterfactual features $\mathcal{FI}(x_S,x_T,W)$ and $\mathcal{FI}(x_T,x_S,W)$, respectively. The $Q_{x_S}$ and $Q_{x_T}$ denote the label prediction certainty and are quantified as $Q_{x}=-e^{\mathcal{H}(p)}$, where $\mathcal{H}(\cdot)$ is the information entropy function and $p$ is the prediction probability of sample $x$. Since the target labels are unavailable, the pseudo-labels are utilized. The certainty-guided loss re-weighting to progressively reduce $F$'s dependency on $V$ is found to stabilize the training.

The domain-invariance mapping $M$ and intervention module are trained in an adversarial manner against each other to prevent the residence of domain-specific $V$ in the learned discriminability. Specifically, the intervention module is optimized to distinguish the domain-sharable $U$ from domain-specific $V$, detect $F$'s dependence on $V$, and generate counterfactual features that maximize the intervention loss in Equation \ref{eq:distance_total}. Meanwhile, the $M$ is trained to guide both the discriminability and domain-invariance learning to focus on $U$ only by minimizing the intervention loss to reduce $V$'s influence on $F$.

\subsection{Interventional Adversarial Training}
\label{sec:discrepancy}
We learn domain-invariant features to bridge the source and target domains by minimizing the distribution divergence between source $F_s$ and target $F_t$ via adversarial training. Specifically, we train an additional domain discriminator network \emph{D} to distinguish $F_s$ from $F_t$. The negative cross-entropy loss is used to measure how well the learned representation confuses the domain discriminator and as an indicator of domain discrepancy. To guide the domain-invariance learning to focus on domain-sharable $U$, the adversarial training is performed over the representation of the counterfactual features:
\begin{equation}
\small
\mathcal{D}(S,T) = \mathbb{E}_{x_i,x_j \sim X_S} \log[D(F_{\mathcal{FI}(x_i,x_j,W)})]  \\ +\mathbb{E}_{x_i,x_j \sim X_T} \log[1-D(F_{\mathcal{FI}(x_i,x_j,W)})].
\end{equation}
\noindent Following \cite{ganin2016domain}, we implement a gradient reverse layer (GRL) to reduce domain discrepancy.

\subsection{Theoretical Insights}

In this section, we show how IDA better controls target prediction risk. Following the theory of \cite{ben2010theory}, let $\mathcal{H}$ be the hypothesis class and $F$ be the discriminative feature where adaptation is performed with $S$ and $T$ domains , the target risk is upper-bounded by:
\begin{equation}
\small
\forall h \in \mathcal{H}, \varepsilon_{T}(h(F))\leq \varepsilon_{S}(h(F))+ \frac{1}{2} d_{\mathcal{H} \bigtriangleup \mathcal{H}}(S(F),T(F))+ \lambda,
\label{eq:theory09}
\end{equation}
\noindent where 1) $\varepsilon_{S}(h)$ denotes expected source risk; 2) $d_{\mathcal{H} \bigtriangleup \mathcal{H}}(S(F),T(F))$ measures domain discrepancy over the adapted feature $F$ w.r.t. the hypothesis set $\mathcal{H}$; 3) $\lambda=\min_{h \in \mathcal{H}}\left(\varepsilon_{S}(h(F),f_{S}(F))+\varepsilon_{T}(h(F),f_{T}(F)\right)$ is the optimal joint error on both domains, in which $f_{S}$ and $f_{T}$ are the underlying labeling functions. Because the labeling procedures only depend on $U$, instead of $V$, then $f_{S}(F)=f_{S}(U)$ and $f_{T}(F)=f_{T}(U)$. In the following, we discuss the error bound of DA methods with standard domain-invariance learning, as well as the proposed IDA. Since the first term $\varepsilon_{S}(h)$ is generally minimized to be negligibly small by source labeled data, we mainly consider the second and third term. The definitions of $V_S$, $V_T$, $U_S$, $U_T$, and $F=M(U,V)$ are given above.

\textbf{Standard Domain-invariance Learning.} The $M$ is optimized by both source supervised training and domain-invariance learning. Due to the correlation between $V_S$ and label $Y$, supervised training tends to map $V_S$ as important predictors into $F$ for classification, which is conflicting with domain-invariance learning because $V_s$ and $V_t$ are assumed to be distinct. A generally achieved tradeoff between the two objectives is an arbitrary mixup between ($U_S$,$V_S$) and ($U_T$,$V_T$). If $V$ dominates domain discrepancy and then domain-invariance learning, which is commonly encountered in practice, $V_t$ tends to be transformed by $M$ into $V_S$ for domain-invariance learning and thus only $V_S$ is mapped into $F$. Such a tradeoff satisfies both training objectives. In this case, while the second term is significantly reduced, the third term would be unboundedly enlarged, especially the target $\varepsilon_{T}(h(V_S),f_{T}(U_T))$, as verified in Figure \ref{fig:visualization_error_bound}. This is because that $V_S$ is usually non-informative for target label prediction.

\textbf{IDA Method.} In contrast, IDA eases the above conflicts by explicitly guiding supervised training to focus on $U$ for discriminability. Concretely, with intervention on feature to reduce the correlation between $V_S$ and label $Y$, the source supervised training is guided to avoid M mapping $V_S$ as important predictors to $F$, thus both objectives focus on $U$. Since both labeling functions $f_{S}(U)$ and $f_{T}(U)$ are built on $U$, the third term is expected to be negligibly small while the second term is reduced significantly, as shown in Figure \ref{fig:visualization_error_bound}.

\section{Experiments}
We evaluate the proposed IDA with two settings: 1) typical \emph{one-to-one} unsupervised domain adaptation; 2) domain-agnostic adaptation, a more challenging but practical problem of learning from one labeled source domain and adapting to unknown unlabeled target domains \cite{peng2019domain}.

\subsection{Setup}
\noindent \textbf{Office-31.} This dataset is widely used for visual domain adaptation \cite{saenko2010adapting}. It consists of 4,652 images and 31 categories collected from three different domains: \emph{Amazon} (A) from amazon.com, \emph{Webcam} (W) and \emph{DSLR} (D), taken by web camera and digital SLR camera in different environmental settings, respectively. The \emph{Colored Office-31} is a binary classification task by randomly assigning 15 categories from \emph{ Office-31} with a novel category label ``0'' and the rest 16 categories with label ``1'', which is inspired by \cite{arjovsky2019invariant}.

\noindent \textbf{Office-home.} This is one of the most challenging visual domain adaptation datasets \cite{venkateswara2017deep}, which consists of 15,588 images with 65 categories of everyday objects in office and home settings. There are four significantly dissimilar domains: Art (Ar), Clipart (Cl), Product (Pr) and Real-World (Rw).

\noindent \textbf{Office-Caltech10.} This dataset contains the 10 common categories shared by \emph{Office-31} and \emph{Caltech-256} datasets. Besides the \emph{Amazon}, \emph{DSLR}, and \emph{Webcam}, it includes a novel domain: Caltech (C), which are sampled from the \emph{Caltech-256} dataset.

\begin{table*}[]
 \centering
 \small
\caption{Accuracy ($\%$) on the \emph{Office31} dataset for \emph{one-to-one} unsupervised domain adaptation (AlexNet).}
\label{fig:offfice31}
\resizebox{\textwidth}{!}{
\begin{tabular}{cccccccc}
\toprule
 Method & A$\rightarrow$W & D$\rightarrow$W & W$\rightarrow$D &  A$\rightarrow$D  &  W$\rightarrow$A  &  D$\rightarrow$A    &Avg\\
\midrule
AlexNet      &61.6 $\pm$ 0.4 &95.4 $\pm$ 0.3  &99.0 $\pm$ 0.2   &63.8 $\pm$ 0.5  &49.8 $\pm$ 0.4    &51.1 $\pm$ 0.6 & 70.1  \\
DANN \cite{ganin2016domain}        &73.0 $\pm$ 0.5  &96.4 $\pm$ 0.3   & 99.2 $\pm$ 0.3 &72.3 $\pm$ 0.3 &51.2 $\pm$ 0.5   &53.4 $\pm$ 0.4  & 74.3  \\
ADDA \cite{tzeng2017adversarial}        &73.5 $\pm$ 0.6  &96.2 $\pm$ 0.4   &98.8 $\pm$ 0.4  &71.6 $\pm$ 0.4  &53.5 $\pm$ 0.6    &54.6 $\pm$ 0.5  & 74.7  \\
JAN \cite{long2017deep}        &74.9 $\pm$ 0.3  &96.6 $\pm$ 0.2   &99.5 $\pm$ 0.2 &71.8 $\pm$ 0.2  &55.0 $\pm$ 0.4    &  58.3 $\pm$ 0.3  & 76.0  \\
CDAN-M \cite{long2018conditional}   & 78.3 $\pm$ 0.2   & 97.2 $\pm$ 0.1    & \textbf{100.0} $\pm$ 0.0   & 76.3 $\pm$ 0.1   & 57.3 $\pm$ 0.3     & 57.3 $\pm$ 0.2  &77.7   \\
BUMDA \cite{wen2019bayesian}   & 78.9 $\pm$ 0.4   & 96.9 $\pm$ 0.3    & 99.8 $\pm$ 0.2   & 77.8 $\pm$ 0.3   & 56.6 $\pm$ 0.5     &   57.4 $\pm$ 0.4   &  77.9   \\
MSTN \cite{xie2018learning}          & 80.5 $\pm$ 0.4   & 96.9 $\pm$ 0.1    & 99.9 $\pm$ 0.1   & 74.5 $\pm$ 0.4   & 60.0 $\pm$ 0.6     & 62.5 $\pm$ 0.4   &  79.1   \\
CADA-P \cite{kurmi2019attending}          & 83.4 $\pm$ 0.2   & \textbf{99.8} $\pm$ 0.1    & \textbf{100.0} $\pm$ 0.0   & 80.1 $\pm$ 0.1   & 59.8 $\pm$ 0.2     & 59.5 $\pm$ 0.3& 80.4  \\
DM-ADA \cite{xu2020adversarial}          & 83.9 $\pm$ 0.4   & \textbf{99.8} $\pm$ 0.1    & 99.9 $\pm$ 0.1   & 77.5 $\pm$ 0.2   & 64.0 $\pm$ 0.3     & 64.6 $\pm$ 0.3& 81.6  \\
\midrule
Ours(conv5)     & 78.0 $\pm$ 0.3   & 95.8 $\pm$ 0.3    & 99.0 $\pm$ 0.2   & 76.0 $\pm$ 0.4   & 55.5 $\pm$ 0.3     & 54.5 $\pm$ 0.3  &  76.5   \\
Ours(fc6)        & 83.9 $\pm$ 0.4   & 98.9 $\pm$ 0.1    & \textbf{100.0} $\pm$ 0.0   & \textbf{83.3} $\pm$ 0.3   & 63.9 $\pm$ 0.2     & 64.3 $\pm$ 0.3   &  82.4   \\
\textbf{Ours(fc7)}  & \textbf{84.3} $\pm$ 0.4   & 99.3 $\pm$ 0.1    & \textbf{100.0} $\pm$ 0.0    & 82.6 $\pm$ 0.3 & \textbf{64.3} $\pm$ 0.4 & \textbf{64.9} $\pm$ 0.3 & \textbf{82.6} \\
\midrule
Ours(fc7, L2P)        & 82.9 $\pm$ 0.3   & 98.2 $\pm$ 0.4    & 99.8 $\pm$ 0.2   & 82.0 $\pm$ 0.3   & 62.3 $\pm$ 0.2     & 63.0 $\pm$ 0.3   &  81.4   \\
Ours(fc7, L1P)        & 83.7 $\pm$ 0.2   & 98.8 $\pm$ 0.2    & \textbf{100.0} $\pm$ 0.0    & 82.6 $\pm$ 0.3   & 63.7 $\pm$ 0.3     & 64.1 $\pm$ 0.2   &  82.2   \\
\bottomrule
\end{tabular}}
\end{table*}

\setlength{\tabcolsep}{1mm}
\begin{table*}[ht]
 \centering
 \small
\caption{Accuracy ( \% ) on the \emph{Office-home} dataset for \emph{one-to-one} unsupervised domain adaptation (AlexNet).}
\label{tab:office-home}
\resizebox{\textwidth}{!}{
\begin{tabular}{cccccccccccccc}
\toprule
    Method &Ar:Cl&Ar:Pr&Ar:Rw&Cl:Ar&Cl:Pr&Cl:Rw&Pr:Ar&Pr:Cl&Pr:Rw&Rw:Ar &Rw:Cl&Rw:Pr&Avg\\
    \midrule
AlexNet   & 26.3 & 32.6 & 41.3 & 22.1 & 41.7 & 42.1 & 20.5 & 20.3 & 51.1 & 31.0 & 27.9 & 54.9 & 34.3 \\
DANN \cite{ganin2016domain}              & 36.4 & 45.2 & 54.7 & 35.2 & 51.8 & 55.1 & 31.6 & 39.7 & 59.3 & 45.7 & 46.4 & 65.9 & 47.3 \\
JAN \cite{long2017deep}              & 35.5 & 46.1 & 57.7 & 36.4 & 53.3 & 54.5 & 33.4 & 40.3 & 60.1 & 45.9 & 47.4 & 67.9 & 48.2 \\
MSTN \cite{xie2018learning}       & 34.9 & 46.2 & 56.8 & 36.6 & 55.0 & 55.4 & 33.3 & 41.7 & 60.7 & 47.0 & 45.9 & 68.3 & 48.5 \\
CDAN-M \cite{long2018conditional}       & 38.1 & 50.3 & 60.3 & 39.7 & 56.4 & 57.8 & 35.5 & 43.1 & 63.2 & 48.4 & 48.5 & 71.1 & 51.0 \\
GCAN \cite{ma2019gcan}       & 36.4 & 47.3 & 61.1 & 37.9 & 58.3 & 57.0 & 35.8 & 42.7 & 64.5 & 50.1 & 49.1 & 72.5 & 51.1 \\
BUMDA \cite{wen2019bayesian}  & 40.3 & 51.6 & 61.5 & 37.9 & 58.0 & 58.6 & 33.6 & 45.9 & 61.8 & 50.1 & 50.9 & 71.7 & 51.8 \\
ALDA \cite{chen2020adversarial}  & 41.2 & 51.8 & 62.0 & 38.5 & 57.7 & 59.2 & 34.9 & 46.4 & 61.4 & 50.8 & 51.2 & 72.3 & 52.3 \\
\midrule
Ours(conv5)       & 38.2 & 50.0 & 60.3 & 36.5 & 58.1 & 58.3 & 34.4 & 44.2 & 61.7 & 48.8 & 48.7 & 70.2 & 50.8 \\
Ours(fc6)       & 42.9 & \textbf{54.3} & 64.3 & 40.9 & 59.6 & 59.0 & 35.8 & 47.1 & 64.5 & 50.9 & \textbf{52.7} & \textbf{73.1} & 53.8 \\
\textbf{Ours(fc7)}      & \textbf{43.1} & 54.0 & \textbf{65.1} & \textbf{41.3} & \textbf{59.8} & \textbf{60.5} & \textbf{36.4} & \textbf{47.8} & \textbf{66.5} & \textbf{51.3} & 51.8 & 72.9 & \textbf{54.2} \\
\midrule
Ours(fc7, L2P)      & 41.9 & 52.7 & 63.7 & 41.1 & 58.9 & 60.0 & 35.5 & 47.0 & 65.3 & 50.2 & 50.9 & 71.8 & 53.3 \\
Ours(fc7, L1P)       & 43.0 & 53.5 & 63.9 & 40.9 & 59.0 & 59.9 & 36.1 & 46.9 & 66.1 & 50.8 & 51.8 & 72.5 & 53.7 \\
    \bottomrule
\end{tabular}}
\end{table*}
\setlength{\tabcolsep}{3.5pt}

\paragraph{Implementation Details}
We finetune the AlexNet pre-trained from the ImageNet. Following the DANN \cite{ganin2016domain}, a adaptation layer \emph{fcb} with 256 units is added after the \emph{fc7} layer, where the feature intervention is performed. The attention module of the intervention operation consists of two fully-connected layers with a 256-units hidden layer and 4096-units output layer. We set $q=2$ for the interventional feature distance and the weight of intervention loss $\beta =0.1*\gamma$. We set the distance threshold $t_d$ to be 9. We progressively increase the importance of the domain discrepancy and set $\gamma= \frac{2}{1+\exp(k \cdot m)}-1$, where $k=-10$ and $m$ denotes the training progress ranging from 0 to 1.  We use a similar hyper-parameter selection strategy as in DANN. All reported results are averaged from 5 runs.

\subsection{\emph{One-to-one} Unsupervised Domain Adaptation}
Experimenting on the \emph{Office31} and the \emph{Office-home} datasets, the results are reported in Table \ref{fig:offfice31} and Table \ref{tab:office-home}, respectively. The proposed \emph{Ours(fc7)} outperforms the compared methods consistently. On the \emph{Office31} data, the advantages the proposed method over the compared are more distinct when the plain \emph{Webcam} or \emph{DSLR} domains act as source domain, in which case the learned discriminative features tend to overfit source labels and be unsafely transferable. This is also verified by the distinct performance improvements of IDA on the \emph{Office-home} tasks with \emph{Real-world} being the target domain, in which case diversified target features alleviate source-specific correlations through feature intervention and improve transferability of the learned features.

We explore where to perform the feature intervention and experiment on the \emph{conv5}, \emph{fc6} and \emph{fc7} layers, respectively. As shown in Table \ref{fig:offfice31} and Table \ref{tab:office-home}, the performance improves as the intervention layer gets closer to the adaptation layer.

We further study different strategies to achieve the mapping consistency in Equation \ref{eq:distance_total} to alleviate the spurious correlation. We experiment on the predicted probability with \emph{KL}-divergence, L1-distance, and L2-distance. We find that the \emph{KL}-divergence tends to cause unstable training and performing the consistency directly on the adaptation layer, which is proposed, achieves the best performance. Because, in this case, we are able to jointly optimize the discriminability and domain-invariance learning in the same feature layer and guide both to focus on the domain-sharable part.

\begin{table*}[]
\centering
 \small
\caption{Accuracy ( \% ) on the \emph{Office31} dataset for domain-agnostic adaptation (AlexNet).}
 \label{tab:domain_agnostic}
\begin{tabular}{cccccc}
\toprule
 Method & A $\rightarrow$ C,D,W  &  C $\rightarrow$ A,D,W  &  D $\rightarrow$ A,C,W  &  W $\rightarrow$ A,C,D    &Avg\\
\midrule
AlexNet     & 83.1 $\pm$ 0.2    & 88.9 $\pm$ 0.4   & 86.7 $\pm$ 0.4     & 82.2 $\pm$ 0.3  &  85.2   \\
RTN \cite{long2016unsupervised}         & 85.2 $\pm$ 0.4    & 89.8 $\pm$ 0.3   & 81.7 $\pm$ 0.3     & 83.7 $\pm$ 0.4   &  85.1   \\
JAN \cite{long2017deep}        & 83.5 $\pm$ 0.3    & 88.5 $\pm$ 0.2   & 80.1 $\pm$ 0.3     & 85.9 $\pm$ 0.4   &  84.5   \\
DANN \cite{ganin2016domain}        & 84.8 $\pm$ 0.3    & 89.7 $\pm$ 0.2   & 87.8 $\pm$ 0.4     & 89.1 $\pm$ 0.4   &  87.9   \\
DADA \cite{peng2019domain}          & 86.3 $\pm$ 0.3    & 91.7 $\pm$ 0.4  &  89.9 $\pm$ 0.3     & 91.3 $\pm$ 0.3   &  89.8   \\
\midrule
Ours(fc6)          & \textbf{89.0} $\pm$ 0.2    & 92.5 $\pm$ 0.3  &  90.7 $\pm$ 0.2     & 92.4 $\pm$ 0.3   &  91.2   \\
\textbf{Ours(fc7)} & 88.9 $\pm$ 0.3 & \textbf{92.8} $\pm$ 0.3 & \textbf{91.0} $\pm$ 0.3 & \textbf{92.7} $\pm$ 0.2 & \textbf{91.4} \\
\midrule
Ours(fc7, L2P)           & 88.4 $\pm$ 0.2    & 91.7 $\pm$ 0.4  &  90.0 $\pm$ 0.3     & 92.2 $\pm$ 0.4   &  90.8   \\
Ours(fc7, L1P)           & 88.9 $\pm$ 0.3    & 92.7 $\pm$ 0.2  &  90.7 $\pm$ 0.4    & 92.6 $\pm$ 0.3   &  91.2   \\
\bottomrule
\end{tabular}
\end{table*}

\subsection{Domain-agnostic Adaptation}
We evaluate the transferability of the learned discriminative features on the domain-agnostic learning task of adaptation from one labeled source domain to unknown unlabeled target domains with the \emph{Office-Caltech10} dataset. As shown in Table \ref{tab:domain_agnostic}, the proposed \emph{Ours(fc7)} achieves the best performance on most of the tasks. \emph{Negative transfer} \cite{pan2010survey} occurs on RTN and JAN. DADA is specifically designed for this task to explicitly disentangle features. The proposed IDA outperforms DADA by distinguishing domain-sharable features from domain-specifics through feature intervention and jointly guiding both domain-invariance and discriminability learning to focus on the domain-sharable part.

\subsection{Analyses}

\subsubsection{Target Error Bound}
We study the target-domain error bound on two representative tasks: \emph{A$\rightarrow$W} using the \emph{Office-31}
and \emph{W$\rightarrow$A,D,C} using the \emph{Office-Caltech10}. As formulated in Equation \ref{eq:theory09}, the first term is generally ignorable, therefore, only the second term, domain discrepancy, and the third term, joint error of ideal hypothesis, are analyzed.

\begin{figure*}[ht]
  \centering
  \subfigure[\emph{A$\rightarrow$W}]{\includegraphics[scale=0.2]{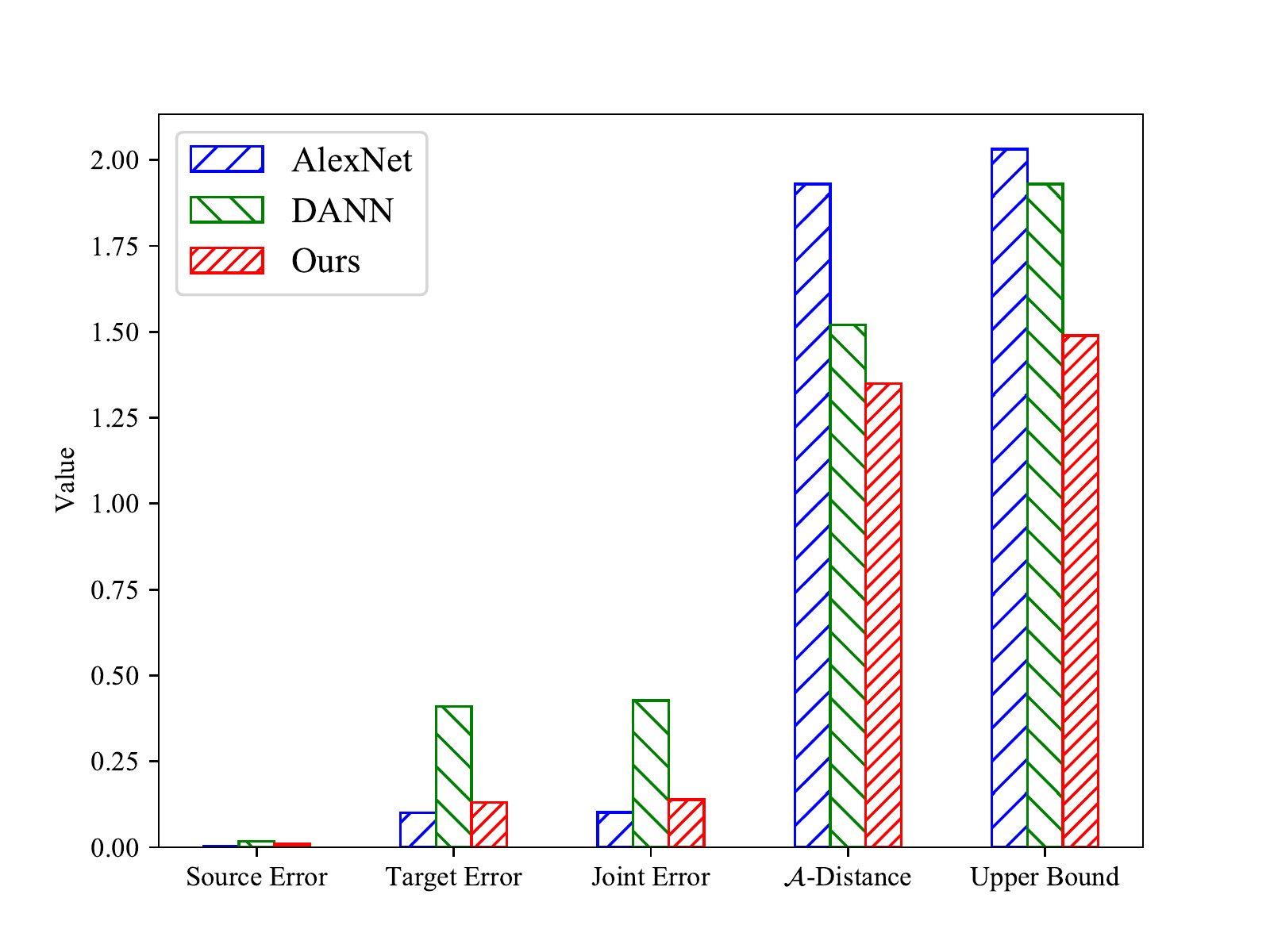}}
  \subfigure[\emph{W$\rightarrow$A,D,C}]{\includegraphics[scale=0.2]{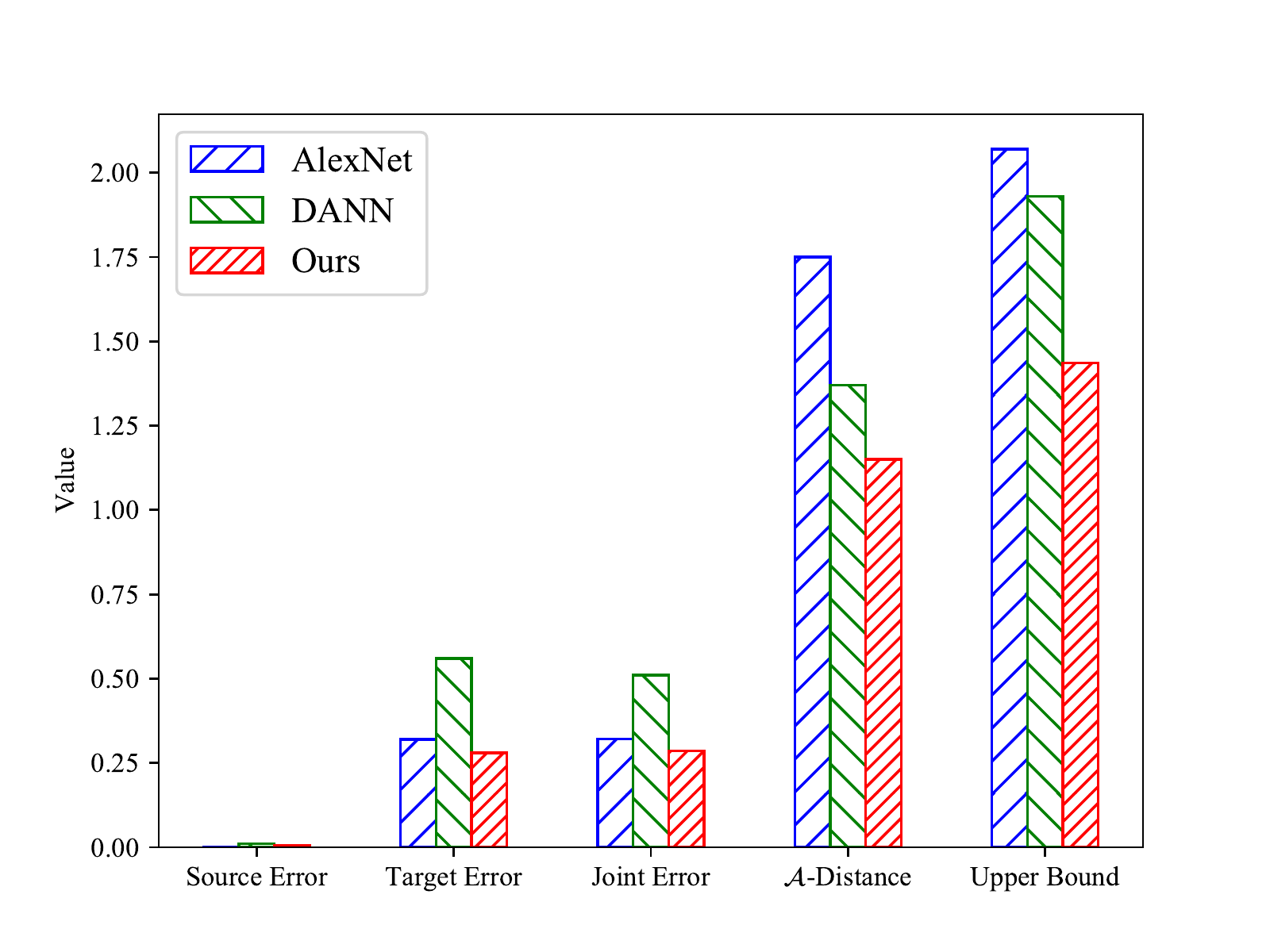}}
  \subfigure[AlexNet]{\includegraphics[scale=0.155]{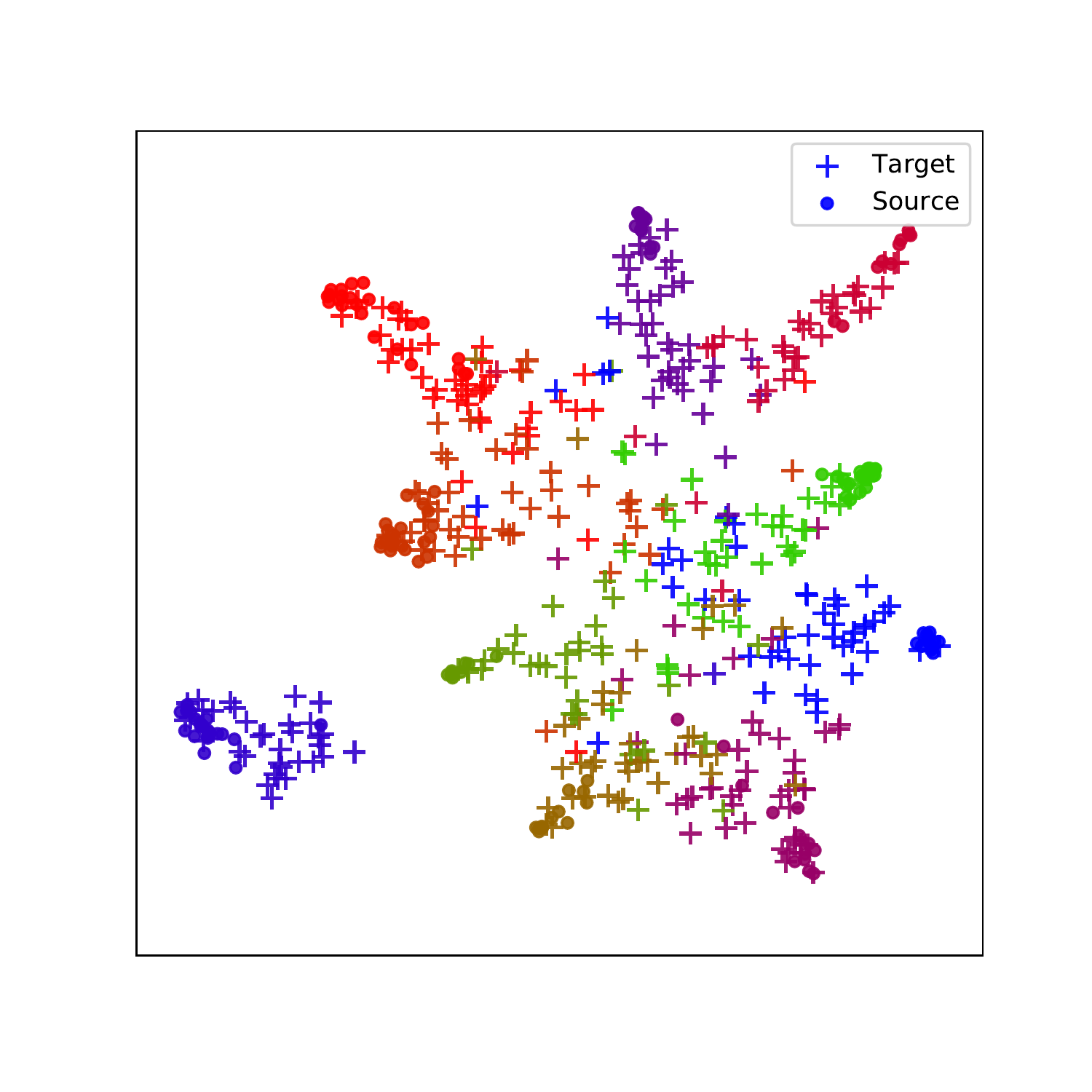}}
  \subfigure[DANN]{\includegraphics[scale=0.155]{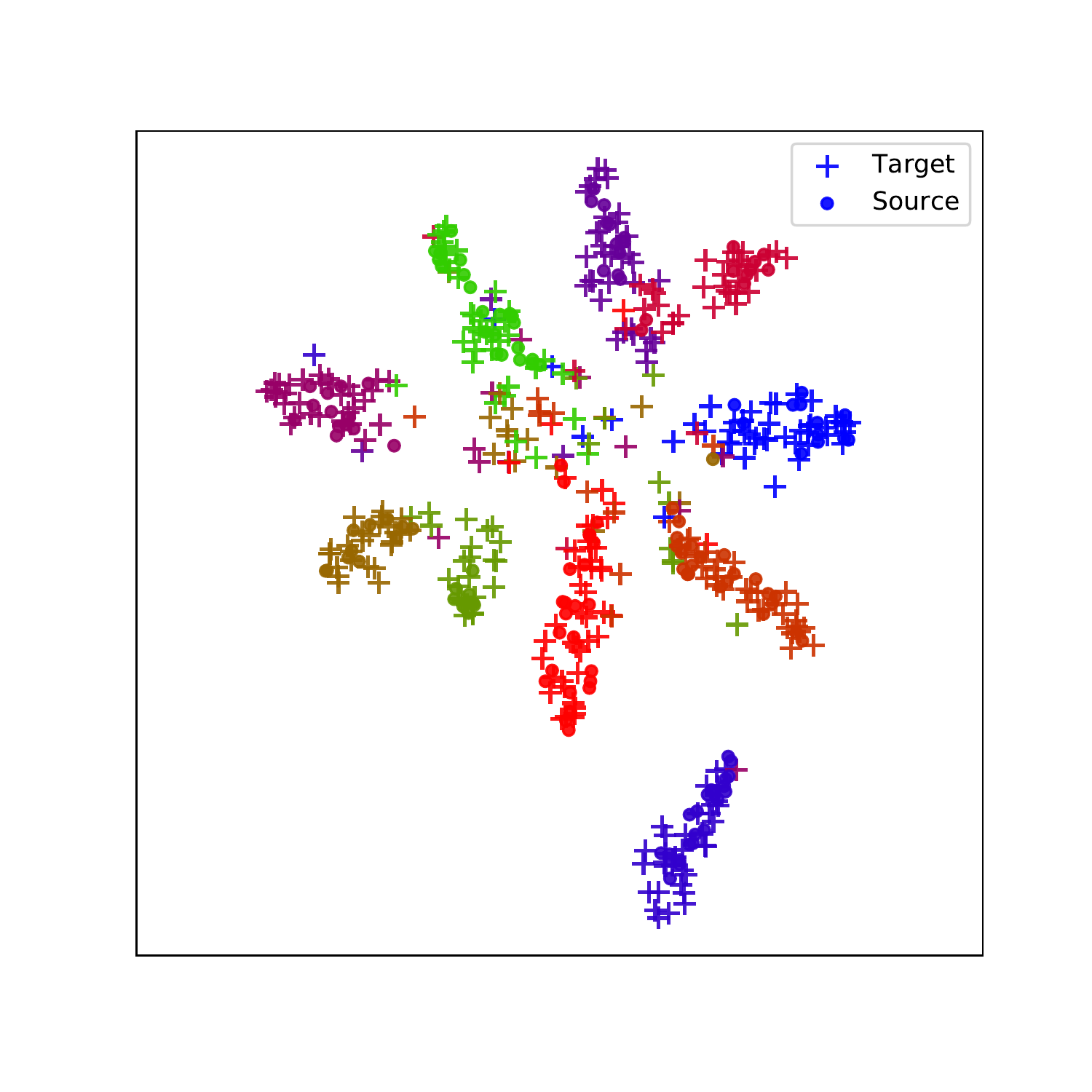}}
  \subfigure[Ours]{\includegraphics[scale=0.155]{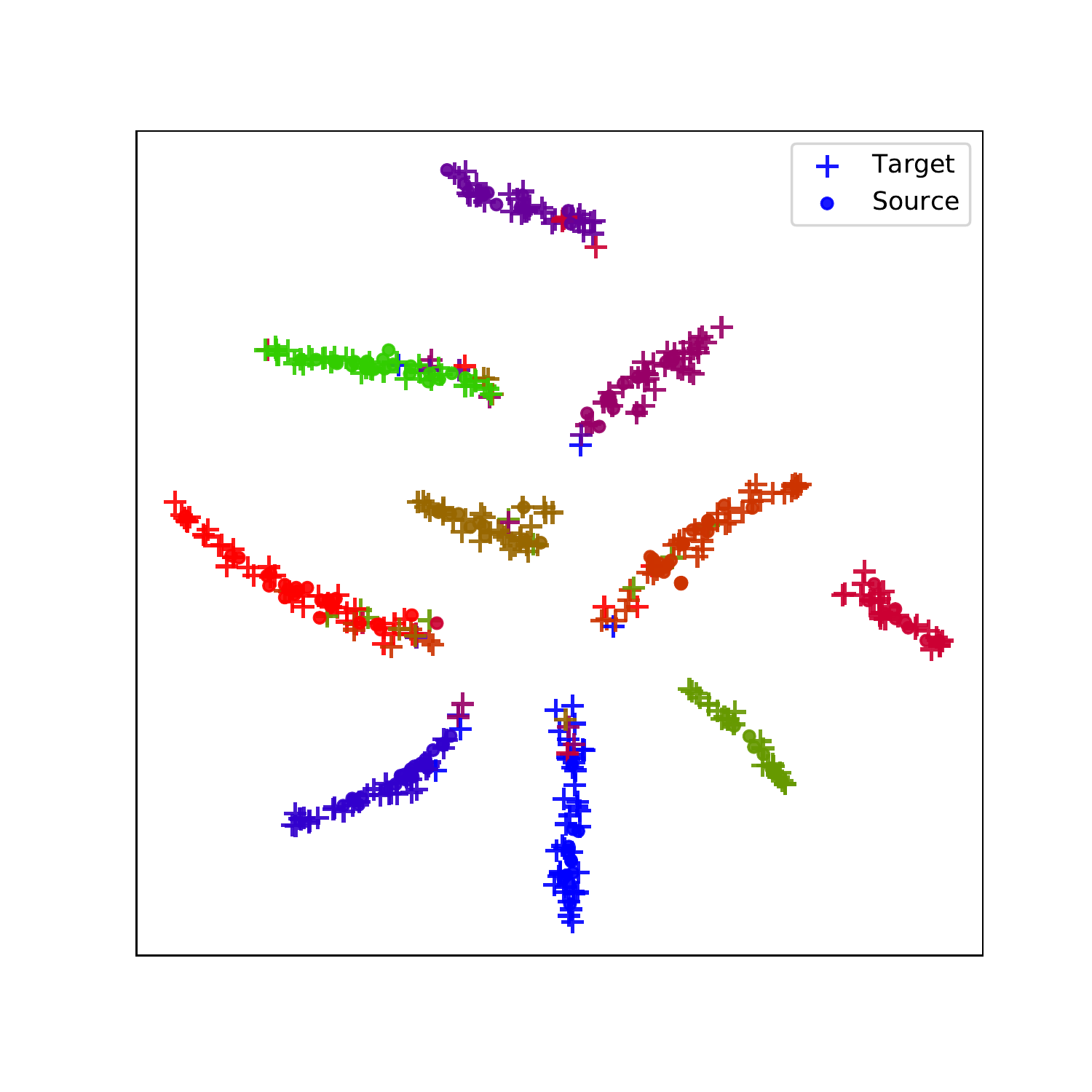}}
  \caption{Target error bound analyses with (a) for \emph{A$\rightarrow$W} and (b) \emph{W$\rightarrow$A,D,C}, and t-SNE visualization of features by (c) AlexNet, (d) DANN, and (e) Ours (different colors denote different category labels).  }
  \label{fig:visualization_error_bound}
\end{figure*}

\noindent \textbf{Domain Discrepancy.} As suggested in \cite{ben2010theory}, domain discrepancy is measured using $\mathcal{A}$-distance, defined as $d_A=2(1-2\epsilon )$, where $\epsilon$ is the generalization error of a domain classifier trained to distinguish source and target features. As shown in Figure \ref{fig:visualization_error_bound}, our method reduces domain discrepancy over fine-tuned AlexNet more significantly than DANN on both tasks.

\noindent \textbf{Error of Ideal Joint Hypothesis.} To evaluate the third term, we train a two-layer MLP classifier on the adapted features from both source and target, using their category labels, as done in \cite{chen2019transferability}. Higher classification accuracy indicates better feature discriminability. As shown in Figure \ref{fig:visualization_error_bound}, DANN deteriorates feature discriminability with enlarged joint error. Our method enlarges the joint error negligibly on the \emph{A$\rightarrow$W} task while reduces it on the \emph{W$\rightarrow$A,D,C} task which indicates our method contributes to learning more discriminative features than using source labels only.

Combining the two terms above, we show that the proposed method reduces domain discrepancy significantly while keeping feature discriminability negligibly altered, thus successfully reducing the upper bound of target error.

\subsubsection{Feature Visualization}
To dive deeper into the learned discriminative features, we visualize them of both domains on the \emph{D$\rightarrow${A,W,C}} task of \emph{Office-Caltech10} using the t-SNE embedding \cite{maaten2008visualizing}. In Figure \ref{fig:visualization_error_bound}, we visualize features of non-adapted models, DANN and the proposed IDA. Comparing to DANN, our method achieves better feature discriminability in target domain by keeping each category well separated while with much less false feature alignment.

\subsubsection{Domain Generalization}
We study domain generalization ability of learned features on the \emph{Office-Caltech10} dataset. We perform adaptation from one source to one target domain, and evaluate learned features in novel unseen domains. We experiment with four representative tasks, with results reported in Table \ref{tab:feature_generalization}. DANN improves feature transferability in the target domain but deteriorates the generalization ability over the AlexNet. The proposed IDA simultaneously enhances the feature transferability to the target domain and novel domains.

\begin{table*}
\centering
 \small
\caption{Accuracy ($\%$) on the \emph{Office-Caltech10} dataset for domain generalization (AlexNet). }
\label{tab:feature_generalization}
    \begin{tabular}{ccccccccccc}
    \toprule
    \multirow{2}{*}{Method}&
    \multicolumn{2}{c}{ A$\rightarrow$W}&\multicolumn{2}{c}{W$\rightarrow$D}&\multicolumn{2}{c}{D$\rightarrow$A}&\multicolumn{2}{c}{C$\rightarrow$A}&\multicolumn{2}{c}{Avg}\\
    \cmidrule(lr){2-3} \cmidrule(lr){4-5} \cmidrule(lr){6-7} \cmidrule(lr){8-9}\cmidrule(lr){10-11}
    &W&C,D  &D&A,C  &A&C,W   &A&D,W   &Target&Unseen\\
    \midrule
   AlexNet  & 75.0 & 86.2 & 98.6 & 69.8 & 83.6 & 86.1 & 91.7 & 85.0 & 87.2 & 81.8 \\
    DANN \cite{ganin2016domain} & 88.7 & 85.5 & 99.2 & 66.2 & 87.3 & 81.1 & 93.8 & 87.1 & 92.3 & 80.0 \\
    \midrule
        Ours(fc6) & \textbf{94.1} & 91.9 & 99.8 & 77.7 & 93.0 & 88.9 & \textbf{96.0} & \textbf{89.1} & 95.7 & 86.9 \\
    \textbf{Ours(fc7)} & 93.9 & \textbf{92.9} & \textbf{100.0} & \textbf{78.7} & \textbf{93.7} & \textbf{89.5} & 95.7 & 88.9 & \textbf{95.8} & \textbf{87.5} \\
    \midrule
    Ours(fc7, L2P) & 93.0 & 91.9 & \textbf{100.0} & 76.9 & 92.8 & 88.9 & 94.9 & 87.0 & 95.2 & 86.2 \\
    Ours(fc7, L1P) & 93.7 & 92.3 & \textbf{100.0} & 77.9 & 93.5 & 89.2 & 95.5 & 87.8 & 95.7 & 86.8 \\
    \bottomrule
    \end{tabular}
\end{table*}

\subsubsection{Discriminability Visualization}
\begin{figure*}[tb]
  \centering
  \subfigure[Source domain]{\includegraphics[scale=0.2]{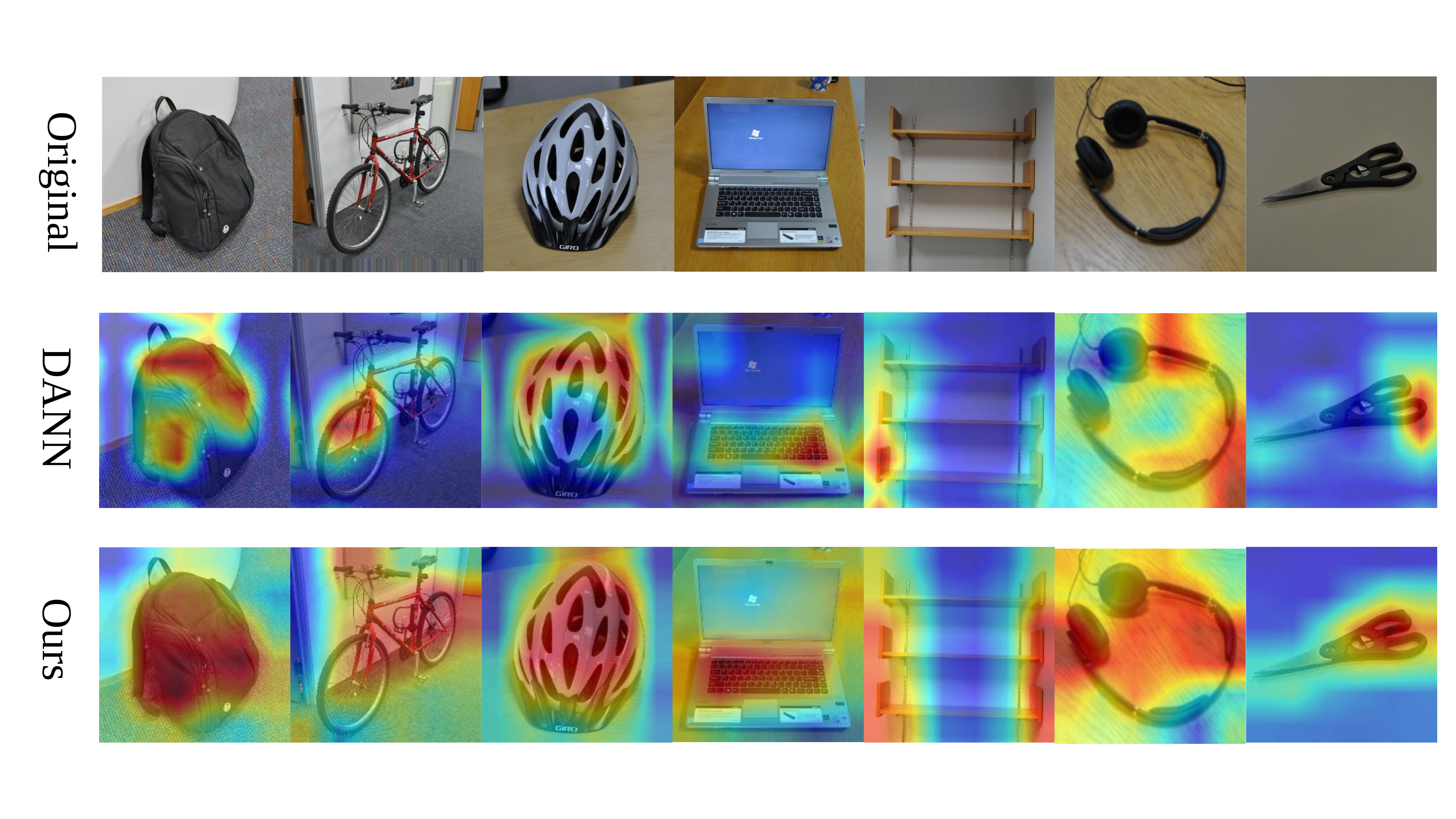}}
  \subfigure[Target domain]{\includegraphics[scale=0.2]{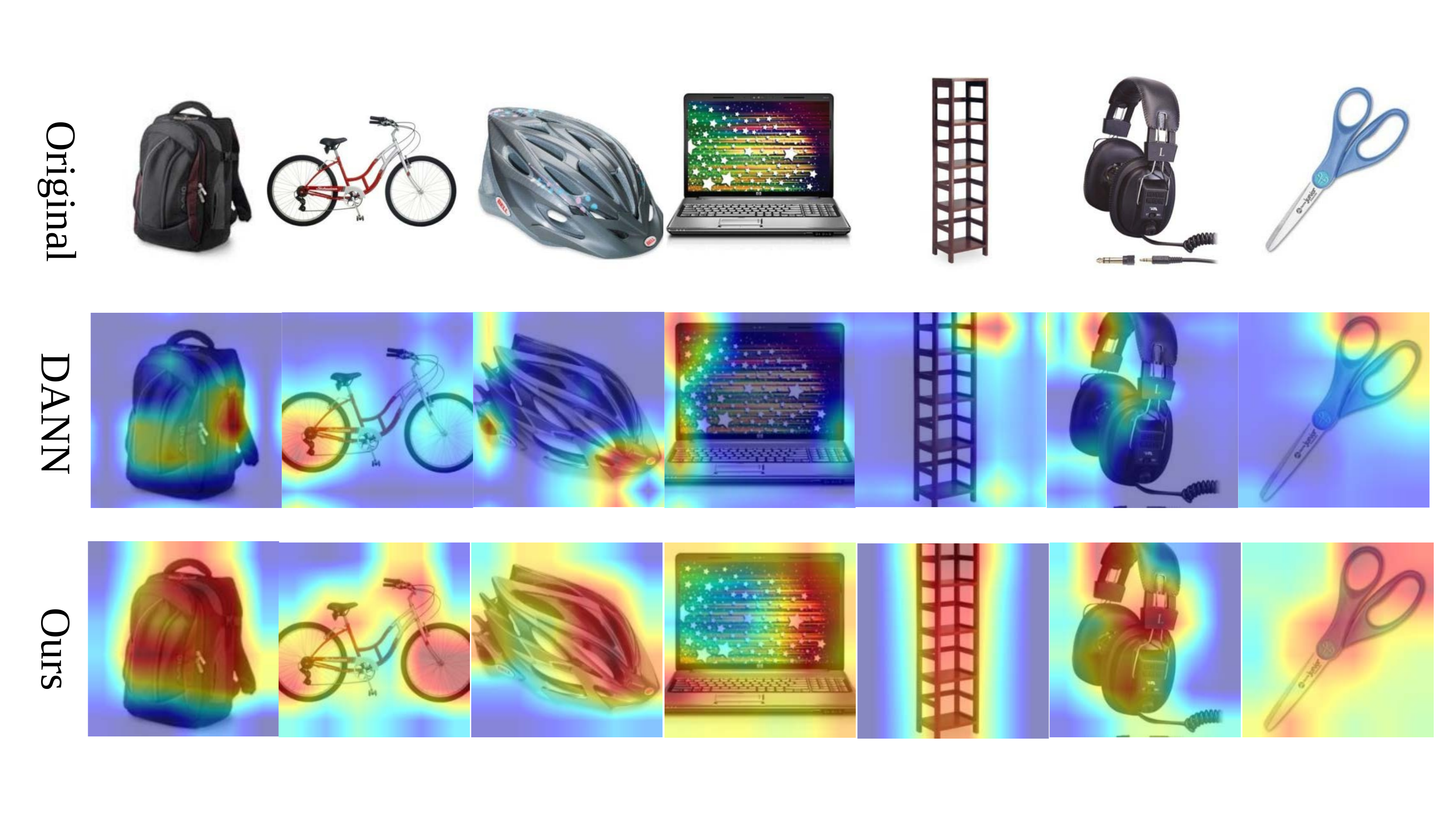}}
  \caption{Discriminability visualization on the \emph{D$\rightarrow$A} task of DANN and the proposed method on the \emph{Office31} dataset. DANN pays attention to trivial features while our method focuses on stably transferable features.}
  \label{fig:visualization}
\end{figure*}

The learned important predictors are highlighted to show the proposed IDA reduces the influence of spurious correlations and learns discriminative feature that are safely transferable. We utilize Grad-CAM \cite{selvaraju2017grad} to visualize the class-discriminative feature map extracted from the last pooling layer in \emph{AlexNet} on the challenging task \emph{D$\rightarrow$A}, with some samples shown in Figure \ref{fig:visualization}. The features learned by DANN tend to focus on some trivial parts of the objects while our method pays more attention to the stably discriminative parts.

\section{Conclusion}

In this paper, we found standard domain invariance learning suffers from spurious correlations and would learn discriminative features that are inferiorly transferable. To address this issue, a novel interventional domain adaptation method was proposed to reduce the correlations and learn more transferable discriminative features. We generated counterfactual features via the proposed feature intervention strategy and intervened in the discriminability learning to guide it to get rid of the domain-specifics and focus on domain-sharable features. The experimental results on both \emph{one-to-one} unsupervised domain adaptation and domain-agnostic adaptation tasks validated that our learned discriminative features are more transferable and generalize well to novel domains.

\section*{Broader Impact}

This domain adaptation method enables more unbiased and reliable data or knowledge utilization. It does not present foreseeable negative societal consequences.



\begin{thebibliography}{10}

\bibitem{csurka2017domain}
Gabriela Csurka.
\newblock {\em Domain adaptation in computer vision applications}, volume~2.
\newblock Springer, 2017.

\bibitem{freitag2016fast}
Markus Freitag and Yaser Al-Onaizan.
\newblock Fast domain adaptation for neural machine translation.
\newblock {\em arXiv preprint arXiv:1612.06897}, 2016.

\bibitem{ben2010theory}
Shai Ben-David, John Blitzer, Koby Crammer, Alex Kulesza, Fernando Pereira, and
  Jennifer~Wortman Vaughan.
\newblock A theory of learning from different domains.
\newblock {\em Machine learning}, 79(1-2):151--175, 2010.

\bibitem{long2015learning}
Mingsheng Long, Yue Cao, Jianmin Wang, and Michael~I Jordan.
\newblock Learning transferable features with deep adaptation networks.
\newblock {\em arXiv preprint arXiv:1502.02791}, 2015.

\bibitem{long2017deep}
Mingsheng Long, Han Zhu, Jianmin Wang, and Michael~I Jordan.
\newblock Deep transfer learning with joint adaptation networks.
\newblock In {\em Proceedings of the 34th International Conference on Machine
  Learning-Volume 70}, pages 2208--2217. JMLR. org, 2017.

\bibitem{peng2019moment}
Xingchao Peng, Qinxun Bai, Xide Xia, Zijun Huang, Kate Saenko, and Bo~Wang.
\newblock Moment matching for multi-source domain adaptation.
\newblock In {\em Proceedings of the IEEE International Conference on Computer
  Vision}, pages 1406--1415, 2019.

\bibitem{ganin2016domain}
Yaroslav Ganin, Evgeniya Ustinova, Hana Ajakan, Pascal Germain, Hugo
  Larochelle, Fran{\c{c}}ois Laviolette, Mario Marchand, and Victor Lempitsky.
\newblock Domain-adversarial training of neural networks.
\newblock {\em The Journal of Machine Learning Research}, 17(1):2096--2030,
  2016.

\bibitem{tzeng2017adversarial}
Eric Tzeng, Judy Hoffman, Kate Saenko, and Trevor Darrell.
\newblock Adversarial discriminative domain adaptation.
\newblock In {\em Proceedings of the IEEE Conference on Computer Vision and
  Pattern Recognition}, pages 7167--7176, 2017.

\bibitem{rozantsev2018beyond}
Artem Rozantsev, Mathieu Salzmann, and Pascal Fua.
\newblock Beyond sharing weights for deep domain adaptation.
\newblock {\em IEEE transactions on pattern analysis and machine intelligence},
  41(4):801--814, 2018.

\bibitem{chen2019transferability}
Xinyang Chen, Sinan Wang, Mingsheng Long, and Jianmin Wang.
\newblock Transferability vs. discriminability: Batch spectral penalization for
  adversarial domain adaptation.
\newblock In {\em International Conference on Machine Learning}, pages
  1081--1090, 2019.

\bibitem{arjovsky2019invariant}
Martin Arjovsky, L{\'e}on Bottou, Ishaan Gulrajani, and David Lopez-Paz.
\newblock Invariant risk minimization.
\newblock {\em arXiv preprint arXiv:1907.02893}, 2019.

\bibitem{saenko2010adapting}
Kate Saenko, Brian Kulis, Mario Fritz, and Trevor Darrell.
\newblock Adapting visual category models to new domains.
\newblock In {\em European conference on computer vision}, pages 213--226.
  Springer, 2010.

\bibitem{venkateswara2017deep}
Hemanth Venkateswara, Jose Eusebio, Shayok Chakraborty, and Sethuraman
  Panchanathan.
\newblock Deep hashing network for unsupervised domain adaptation.
\newblock In {\em Proc. CVPR}, pages 5018--5027, 2017.

\bibitem{zhao2019learning}
Han Zhao, Remi~Tachet Des~Combes, Kun Zhang, and Geoffrey Gordon.
\newblock On learning invariant representations for domain adaptation.
\newblock In {\em International Conference on Machine Learning}, pages
  7523--7532, 2019.

\bibitem{xie2018learning}
Shaoan Xie, Zibin Zheng, Liang Chen, and Chuan Chen.
\newblock Learning semantic representations for unsupervised domain adaptation.
\newblock In {\em International Conference on Machine Learning}, pages
  5423--5432, 2018.

\bibitem{pei2018multi}
Zhongyi Pei, Zhangjie Cao, Mingsheng Long, and Jianmin Wang.
\newblock Multi-adversarial domain adaptation.
\newblock In {\em AAAI Conference on Artificial Intelligence}, pages
  3934--3941, 2018.

\bibitem{saito2018maximum}
Kuniaki Saito, Kohei Watanabe, Yoshitaka Ushiku, and Tatsuya Harada.
\newblock Maximum classifier discrepancy for unsupervised domain adaptation.
\newblock In {\em Proceedings of the IEEE Conference on Computer Vision and
  Pattern Recognition}, pages 3723--3732, 2018.

\bibitem{xu2020adversarial}
Minghao Xu, Jian Zhang, Bingbing Ni, Teng Li, Chengjie Wang, Qi~Tian, and
  Wenjun Zhang.
\newblock Adversarial domain adaptation with domain mixup.
\newblock In {\em The Thirty-Fourth AAAI Conference on Artificial
  Intelligence}, pages 6502--6509. AAAI Press, 2020.

\bibitem{liu2019transferable}
Hong Liu, Mingsheng Long, Jianmin Wang, and Michael Jordan.
\newblock Transferable adversarial training: A general approach to adapting
  deep classifiers.
\newblock In {\em International Conference on Machine Learning}, pages
  4013--4022, 2019.

\bibitem{kurmi2019attending}
Vinod~Kumar Kurmi, Shanu Kumar, and Vinay~P Namboodiri.
\newblock Attending to discriminative certainty for domain adaptation.
\newblock In {\em Proceedings of the IEEE Conference on Computer Vision and
  Pattern Recognition}, pages 491--500, 2019.

\bibitem{chen2020adversarial}
Minghao Chen, Shuai Zhao, Haifeng Liu, and Deng Cai.
\newblock Adversarial-learned loss for domain adaptation.
\newblock In {\em AAAI}, pages 3521--3528, 2020.

\bibitem{hu2018squeeze}
Jie Hu, Li~Shen, and Gang Sun.
\newblock Squeeze-and-excitation networks.
\newblock In {\em Proceedings of the IEEE conference on computer vision and
  pattern recognition}, pages 7132--7141, 2018.

\bibitem{peng2019domain}
Xingchao Peng, Zijun Huang, Ximeng Sun, and Kate Saenko.
\newblock Domain agnostic learning with disentangled representations.
\newblock In {\em International Conference on Machine Learning}, pages
  5102--5112, 2019.

\bibitem{long2018conditional}
Mingsheng Long, ZHANGJIE CAO, Jianmin Wang, and Michael~I Jordan.
\newblock Conditional adversarial domain adaptation.
\newblock In {\em Advances in Neural Information Processing Systems 31}, pages
  1647--1657. Curran Associates, Inc., 2018.

\bibitem{wen2019bayesian}
Jun Wen, Nenggan Zheng, Junsong Yuan, Zhefeng Gong, and Changyou Chen.
\newblock Bayesian uncertainty matching for unsupervised domain adaptation.
\newblock In {\em Proceedings of the 28th International Joint Conference on
  Artificial Intelligence}, pages 3849--3855. AAAI Press, 2019.

\bibitem{ma2019gcan}
Xinhong Ma, Tianzhu Zhang, and Changsheng Xu.
\newblock Gcan: Graph convolutional adversarial network for unsupervised domain
  adaptation.
\newblock In {\em Proceedings of the IEEE Conference on Computer Vision and
  Pattern Recognition}, pages 8266--8276, 2019.

\bibitem{long2016unsupervised}
Mingsheng Long, Han Zhu, Jianmin Wang, and Michael~I Jordan.
\newblock Unsupervised domain adaptation with residual transfer networks.
\newblock In {\em Advances in neural information processing systems}, pages
  136--144, 2016.

\bibitem{pan2010survey}
Sinno~Jialin Pan, Qiang Yang, et~al.
\newblock A survey on transfer learning.
\newblock {\em IEEE Transactions on knowledge and data engineering},
  22(10):1345--1359, 2010.

\bibitem{maaten2008visualizing}
Laurens van~der Maaten and Geoffrey Hinton.
\newblock Visualizing data using t-sne.
\newblock {\em Journal of machine learning research}, 9(Nov):2579--2605, 2008.

\bibitem{selvaraju2017grad}
Ramprasaath~R Selvaraju, Michael Cogswell, Abhishek Das, Ramakrishna Vedantam,
  Devi Parikh, and Dhruv Batra.
\newblock Grad-cam: Visual explanations from deep networks via gradient-based
  localization.
\newblock In {\em Proceedings of the IEEE international conference on computer
  vision}, pages 618--626, 2017.

\end{thebibliography}
\end{document}